\documentstyle[mjmacros,tree-dvips,lfg97,ipa,epic,eepic,fullname,palatino,11pt]{article}

\title{Features as Resources in R-LFG}
\author{Mark Johnson\\ Brown University}

\newcommand{\ul}[1]{\underline{#1}}

\begin{document} \bibliographystyle{fullname} 

\maketitle

\section{Introduction}

\noindent
This paper describes a new formalization of Lexical-Functional Grammar
called R-LFG (where the ``R'' stands for ``Resource-based'').  The
formal details of R-LFG are presented in \namecite{Johnson97}; the
present work concentrates on motivating R-LFG and explaining to
linguists how it differs from the ``classical'' LFG framework
presented in \namecite{Kaplan82}.

This work is largely a reaction to the linear logic semantics for LFG
developed by Dalrymple and colleagues 
\cite{Dalrymple95Lamping,Dalrymple96Saraswat,Dalrymple96bSaraswat,Dalrymple96cSaraswat}.  
As they
note, LFG's f-structure completeness and coherence constraints fall
out as a by-product of the linear logic machinery they propose for
semantic interpretation, thus making those f-structure mechanisms
redundant.  Given that linear logic machinery or something like it is
independently needed for semantic interpretation, it seems reasonable
to explore the extent to which it is capable of handling feature
structure constraints as well.  

R-LFG represents the extreme position that {\em all} linguistically
required feature structure dependencies can be captured by the
resource-accounting machinery of a linear or similiar logic
independently needed for semantic interpretation.  The goal is to show
that LFG linguistic analyses can be expressed as clearly and
perspicuously using the smaller set of mechanisms of R-LFG as they can
using the much larger set of mechanisms in LFG: if this is the case
then we will have shown that positing these extra f-structure
mechanisms are not linguistically warranted.  One way to show this
would be to present a translation procedure which reduces LFGs to
equivalent R-LFGs, but currently no such procedure is known. Thus we
proceed on a case by case basis, demonstrating that particular LFG
analyses can be expressed at least as well in R-LFG.

R-LFG is also of interest because it proposes a radically different
basis for feature structure interaction.  In ``unification-based''
theories of grammar feature structures are typically viewed as static
objects, which are the solutions to systems of feature structure
constraints (called {\em f-descriptions} in LFG) \cite{Kaplan82,Rounds97,Shieber86}.  
However, linguists
often talk informally of ``feature assignment'' and ``feature
checking''; notions which cannot be expressed in a pure unification
grammar.  As discussed below, LFG does contain formal devices which
can expresses these notions indirectly, viz., the non-monotonic
devices such as existential constraints and constraint equations.
On the other hand, the resource oriented nature of R-LFG provides
a direct and natural formalization of the intuitions behind
feature assignment and feature checking.

The rest of this paper is structured as follows.  The next section
sketches the architecture of R-LFG and compares it to that of
standard LFG.  The following section introduces the reader to
the idea that features are resources by demonstrating that one
method of describing agreement relationships in standard LFG 
already possesses a resource-oriented character.  The section
following that describes how very simple agreement relationships
can be described in R-LFG, and the final substantive section
shows how \namecite{Andrews82} analysis of Icelandic Quirky Case 
marking can be re-expressed in R-LFG.

\section{R-LFG: a simplification of LFG}

\noindent
The architectural simplification of R-LFG is best appreciated
when compared with that of standard LFG together with the
linear logic semantics augmentation of Dalrymple et. al.
This section starts by sketching the architecture of standard
LFG, and then presents the revised architecture of R-LFG.

\subsection{The architecture of standard LFG}

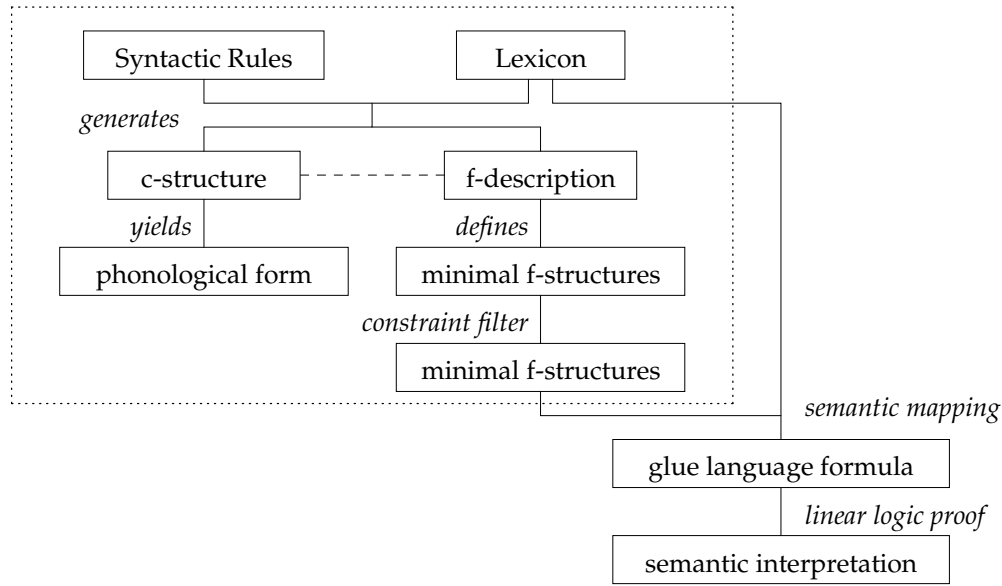
\begin{figure}
\begin{center}\setlength{\unitlength}{0.00083333in}
\begingroup\makeatletter\ifx\SetFigFont\undefined%
\gdef\SetFigFont#1#2#3#4#5{%
  \reset@font\fontsize{#1}{#2pt}%
  \fontfamily{#3}\fontseries{#4}\fontshape{#5}%
  \selectfont}%
\fi\endgroup%
{\renewcommand{\dashlinestretch}{30}
\begin{picture}(6085,3639)(0,-10)
\path(3312,2412)(3312,2112)
\path(2412,2112)(4212,2112)(4212,1812)
	(2412,1812)(2412,2112)
\put(3237,2187){\makebox(0,0)[rb]{\smash{{{\SetFigFont{10}{12.0}{\rmdefault}{\mddefault}{\itdefault}defines}}}}}
\put(3312,1887){\makebox(0,0)[b]{\smash{{{\SetFigFont{10}{12.0}{\familydefault}{\mddefault}{\updefault}minimal f-structures}}}}}
\path(3762,312)(5862,312)(5862,12)
	(3762,12)(3762,312)
\path(4812,612)(4812,312)
\path(3762,912)(5862,912)(5862,612)
	(3762,612)(3762,912)
\put(4812,87){\makebox(0,0)[b]{\smash{{{\SetFigFont{10}{12.0}{\familydefault}{\mddefault}{\updefault}semantic interpretation}}}}}
\put(4812,687){\makebox(0,0)[b]{\smash{{{\SetFigFont{10}{12.0}{\familydefault}{\mddefault}{\updefault}glue language formula}}}}}
\put(4962,1062){\makebox(0,0)[lb]{\smash{{{\SetFigFont{10}{12.0}{\rmdefault}{\mddefault}{\itdefault}semantic mapping}}}}}
\put(4962,387){\makebox(0,0)[lb]{\smash{{{\SetFigFont{10}{12.0}{\rmdefault}{\mddefault}{\itdefault}linear logic proof}}}}}
\path(2262,3012)(2262,2862)(1212,2862)(1212,2712)
\path(2262,2862)(3312,2862)(3312,2712)
\path(2712,2712)(3912,2712)(3912,2412)
	(2712,2412)(2712,2712)
\dashline{60.000}(1812,2562)(2712,2562)
\path(1212,3162)(1212,3012)(3237,3012)(3237,3162)
\path(312,2112)(2112,2112)(2112,1812)
	(312,1812)(312,2112)
\path(1212,2412)(1212,2112)
\path(612,2712)(1812,2712)(1812,2412)
	(612,2412)(612,2712)
\path(2787,3462)(3837,3462)(3837,3162)
	(2787,3162)(2787,3462)
\path(462,3462)(1962,3462)(1962,3162)
	(462,3162)(462,3462)
\path(3312,1812)(3312,1512)
\path(2412,1512)(4212,1512)(4212,1212)
	(2412,1212)(2412,1512)
\path(3312,1212)(3312,1062)(4812,1062)
\path(3387,3162)(3387,3012)(4812,3012)(4812,912)
\put(3312,2487){\makebox(0,0)[b]{\smash{{{\SetFigFont{10}{12.0}{\familydefault}{\mddefault}{\updefault}f-description}}}}}
\put(1212,1887){\makebox(0,0)[b]{\smash{{{\SetFigFont{10}{12.0}{\familydefault}{\mddefault}{\updefault}phonological form}}}}}
\put(1137,2187){\makebox(0,0)[rb]{\smash{{{\SetFigFont{10}{12.0}{\rmdefault}{\mddefault}{\itdefault}yields}}}}}
\put(1212,2487){\makebox(0,0)[b]{\smash{{{\SetFigFont{10}{12.0}{\familydefault}{\mddefault}{\updefault}c-structure}}}}}
\put(3312,3237){\makebox(0,0)[b]{\smash{{{\SetFigFont{10}{12.0}{\familydefault}{\mddefault}{\updefault}Lexicon}}}}}
\put(1212,3237){\makebox(0,0)[b]{\smash{{{\SetFigFont{10}{12.0}{\familydefault}{\mddefault}{\updefault}Syntactic Rules}}}}}
\put(1062,2862){\makebox(0,0)[rb]{\smash{{{\SetFigFont{10}{12.0}{\rmdefault}{\mddefault}{\itdefault}generates}}}}}
\put(3237,1587){\makebox(0,0)[rb]{\smash{{{\SetFigFont{10}{12.0}{\rmdefault}{\mddefault}{\itdefault}constraint filter}}}}}
\put(3312,1287){\makebox(0,0)[b]{\smash{{{\SetFigFont{10}{12.0}{\familydefault}{\mddefault}{\updefault}minimal f-structures}}}}}
\dottedline{45}(12,3612)(4512,3612)(4512,1137)
	(12,1137)(12,3612)
\end{picture}
}\end{center}
\caption{The architecture of standard LFG.  The linear logic semantics
component is shown outside the dotted box. \label{f:lfg1a}}
\end{figure}

\noindent
Figure~\ref{f:lfg1a} shows the architecture of this ``standard'' LFG.
The components of LFG as presented by \namecite{Kaplan82} are
shown inside the dotted box in this figure, and the linear logic
machinery for semantic interpretation posited by Dalrymple et. al.
is depicted outside this box: see these references for further details.

In LFG, a syntactic description of an utterance is taken to be a pair
constiting of a c-structure and an f-structure.\footnote{There are
proposals for additional structures, which for simplicity are ignored
here.}  The yield of the c-structure tree determines the phonological
form of the sentence it describes.

The c-structure/f-structure pairs generated by an LFG are determined
by the following procedure.  The syntactic rules and lexical entries
of an LFG together generate a set of c-structure trees, each of which
is paired with a formula called an f-description which identifies
which (if any) f-structures this c-structure can be paired with.
The f-descriptions are boolean combinations of equations.  These
equations come in two kinds: {\em defining} and {\em constraining}
equations.  

The simplest account of the relationship between f-descriptions and
the f-structures they describe seems to be procedural, following
\namecite{Kaplan82}.\footnote{ See \namecite{Johnson95b} for an
attempt to provide a declarative interpretation for constraining
equations using circumscription, and comments on why such an approach
seems to be impossible in a logic where boolean connectives are
interpreted classically.}  First, the f-description is expanded into
Disjunctive Normal Form (DNF) and the f-structure solution to each
conjunct is determined as follows. The constraining equations are
temporarily ignored (i.e., replaced with {\em true}) and if the
resulting formula is satisfiable and has a unique minimal satisfying
f-structure, that f-structure is a candidate solution to the conjunct.
This candidate solution is a (true) solution to the conjunct just in
case it also satisfies the formula obtained by replacing each
constraining equation in the conjunct with corresponding defining
equations.  The set of solutions to an f-description is the union of
the set of solutions to each conjunct of its DNF, so the f-description
determines a finite number of f-structures.

Dalrymple et.\ al.\ use these f-structures as the input to their
semantic interpretation procedure.  Certain elements in an f-structure
are associated with formula in a {\em glue language}, which is an
amalgam of linear logic and classical first-order logic, in effect
mapping each f-structure into a formula of the glue language For
semantic interpretation to succeed this glue language formula must
derive a term with the type of a saturated proposition: the argument
of this term is the semantic interpretation of the sentence.

\subsection{The architecture of R-LFG}

\begin{figure}
\begin{center}\setlength{\unitlength}{0.00083333in}
\begingroup\makeatletter\ifx\SetFigFont\undefined%
\gdef\SetFigFont#1#2#3#4#5{%
  \reset@font\fontsize{#1}{#2pt}%
  \fontfamily{#3}\fontseries{#4}\fontshape{#5}%
  \selectfont}%
\fi\endgroup%
{\renewcommand{\dashlinestretch}{30}
\begin{picture}(4374,2364)(0,-10)
\path(1062,2037)(1062,1887)(2862,1887)(2862,2037)
\path(312,2337)(1812,2337)(1812,2037)
	(312,2037)(312,2337)
\path(1962,1887)(1962,1737)
\path(1362,1737)(2562,1737)(2562,1437)
	(1362,1437)(1362,1737)
\path(12,1137)(1662,1137)(1662,837)
	(12,837)(12,1137)
\path(2562,1137)(3762,1137)(3762,837)
	(2562,837)(2562,1137)
\path(1962,1437)(1962,1287)
\path(3162,837)(3162,537)
\path(2412,2337)(3462,2337)(3462,2037)
	(2412,2037)(2412,2337)
\path(3012,2037)(3012,1887)(3237,1887)(3237,1137)
\path(762,1137)(762,1287)(3087,1287)(3087,1137)
\path(1962,537)(4362,537)(4362,12)
	(1962,12)(1962,537)
\put(1062,2112){\makebox(0,0)[b]{\smash{{{\SetFigFont{10}{12.0}{\familydefault}{\mddefault}{\updefault}Syntactic Rules}}}}}
\put(1962,1512){\makebox(0,0)[b]{\smash{{{\SetFigFont{10}{12.0}{\familydefault}{\mddefault}{\updefault}c-structure}}}}}
\put(837,912){\makebox(0,0)[b]{\smash{{{\SetFigFont{10}{12.0}{\familydefault}{\mddefault}{\updefault}phonological form}}}}}
\put(3162,912){\makebox(0,0)[b]{\smash{{{\SetFigFont{10}{12.0}{\familydefault}{\mddefault}{\updefault}f-term}}}}}
\put(3162,312){\makebox(0,0)[b]{\smash{{{\SetFigFont{10}{12.0}{\familydefault}{\mddefault}{\updefault}type well-formedness proof}}}}}
\put(3312,612){\makebox(0,0)[lb]{\smash{{{\SetFigFont{10}{12.0}{\rmdefault}{\mddefault}{\itdefault}proof}}}}}
\put(912,1812){\makebox(0,0)[rb]{\smash{{{\SetFigFont{10}{12.0}{\rmdefault}{\mddefault}{\itdefault}generates}}}}}
\put(2937,2112){\makebox(0,0)[b]{\smash{{{\SetFigFont{10}{12.0}{\familydefault}{\mddefault}{\updefault}Lexicon}}}}}
\put(687,1287){\makebox(0,0)[rb]{\smash{{{\SetFigFont{10}{12.0}{\rmdefault}{\mddefault}{\itdefault}labelling}}}}}
\put(3162,87){\makebox(0,0)[b]{\smash{{{\SetFigFont{10}{12.0}{\familydefault}{\mddefault}{\updefault}= semantic interpretation}}}}}
\end{picture}
}\end{center}
\caption{The architecture of R-LFG. \label{f:lfg2}}
\end{figure}
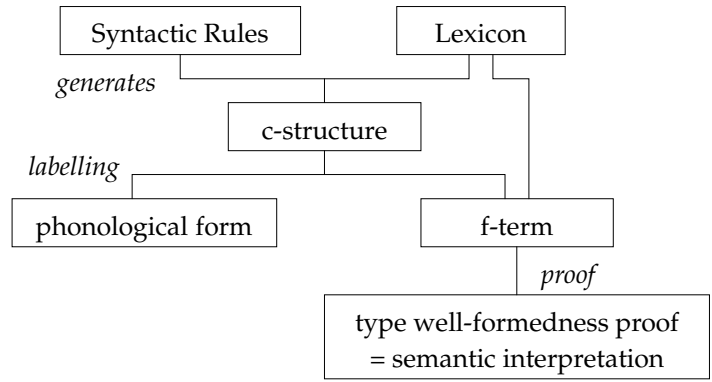

\noindent
The architecture of R-LFG is depicted in Figure~\ref{f:lfg2}.
The most striking difference between LFG and R-LFG is that
R-LFG does not contain an independent level of f-structure
representation, since the same mechanisms used for semantic
interpretation are also used to account for syntactic feature
dependencies.  Given that it is a simpler architecture,
it should be preferred on grounds of parsimony.

The lexical entries and syntactic rules of R-LFG generate
c-structure/f-term pairs in the same way that they generate
c-structure/f-description pairs in LFG.  In LFG several steps are
required to obtain the f-structures that serve as the input to
semantic interpretation from the f-descriptions.  However, in R-LFG
the f-term serves as the input to semantic interpretation directly.
Thus in R-LFG the linguistic effects of f-structure constraints
must be obtained by other means, viz., the same logical mechanisms
used for semantic interpretation.

As explained below, these logical mechanisms enforce a {\em resource
accounting} which ensures that every predicate combines with an
appropriate number of arguments and that every non-root semantic unit 
appears as the argument of some predicate.  The semantic interpretation
itself is determined by the pattern of predicate-argument combination
via the Curry-Howard correspondence, as explained in \namecite{Johnson97}.
Since syntax rather than semantics is the focus of this paper, semantic
interpretation is not discussed further here.

This same resource accounting mechanism is also used to describe
feature dependencies.  Purely syntactic features with no semantic
content differ from semantically interpreted elements only in that
they are semantically vacuous, i.e., given trivial interpretations
which are systematically ignored by any functors which take them as
arguments.

The resource logic used here differs considerably from the glue
language used by Dalrymple et.\ al.  That language includes
first-order terms with equality, which can be used to encode feature
structure unification in the manner of e.g., Definite Clause Grammars
(see \namecite{Shieber86} for a description of the relationship
between the first-order terms of Definite Clause Grammars and
attribute-value ``unification'' grammars) and hence directly
simulate f-structure attribute-value constraints.  While this
would provide a straightforward way to encode f-structure
constraints in the glue language, it is not clear that such
an approach would constitute a real simplification of LFG,
rather than just a reshuffling of its complexity.

For this reason, R-LFG uses a much simpler resource logic than the
glue language of Dalrymple et.\ al.  Inspired by recent work in
Categorial Grammar such as \namecite{Morrill94}, the resource logic is
propositional modal logic that encodes the types of the semantic
objects being manipulated, and the semantic interpretation itself is
provided by a Curry-Howard correspondence between proofs and
$\lambda$-terms \cite{Girard89}.  As \namecite{vanBenthem95}
demonstrates, a wide variety of substructural logics possess a
Curry-Howard correspondence, so the requiremnt that semantic
interpretation is obtained in this way does not identify a particular
logic.  Rather, the precise logic used can be chosen to simplify the
overall linguistic theory.  \namecite{Moortgat:Handbook} develops the
theory of propositional multimodal logics used here.  The reader is
referred to \namecite{Johnson97} for the full details of R-LFG.

\section{Describing agreement relationships with LFG}

\noindent
This section discusses two methods often used for describing agreement
relationships in LFGs.  It turns out that one method, which crucially
relies on ``constraining equations'', can be viewed as describing
agreement in terms of resource dependencies.  Thus resource based
accounts of agreement are not a new innovation of R-LFG, but are
already a familiar part of LFG.  The principal claim behind R-LFG
is that {\em all} linguistic dependencies can be expressed in this
manner, and that the explicit resource-orientation of R-LFG simplifies
and clarifies the nature of the linguistic dependencies concerned.

As sketched above and explained in more detail in \namecite{Kaplan82},
LFG's f-descriptions contain two different kinds of equations.  A
defining equation instantiates the value of an attribute, while a
constraining equation checks that a value is instantiated
by a defining equation elsewhere in the f-description.
The linguistic dependencies involved in simple agreement can
be described using defining equations alone, or by using
a mixture of defining and constraining equations.
This latter method has a natural resource interpretation.

To keep things clear, the two methods for describing agreement
relationships are explained using the same examples (\ref{e:agr1}).
\eenumsentence{\label{e:agr1} 
 \item Sandy snores. \label{e:agr1a}
 \item Professors snore.}  
Both methods of describing agreement relationships require that the
agreeing items (in (\ref{e:agr1a}), \phon{Sandy} and \phon{snores})
are capable of constraining the value of the same f-structure element;
this is usually achieved by defining equations associated with
syntactic rules.  The agreeing items both impose constraints the value
of that f-structure element, thus ensuring that only compatible items
can appear simultaneously in a syntactic structure.

\subsection{Agreement using defining equations alone}

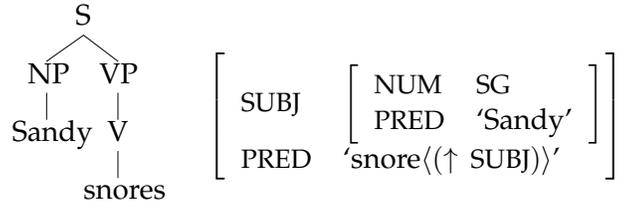
\begin{figure}
\begin{center}
\mbox{\lower3em\hbox{
\begin{picture}(54,78)(0,-78)
%
\put(24,-8){S}
\drawline(27,-12)(13,-22)
\put(6,-30){NP}
\drawline(13,-34)(13,-44)
\put(0,-52){Sandy}
\drawline(27,-12)(40,-22)
\put(33,-30){VP}
\drawline(40,-34)(40,-44)
\put(36,-52){V}
\drawline(40,-56)(40,-66)
\put(27,-74){snores}
\end{picture}}}
\hspace{0.2in}
\fs{ \SUBJ & \tfs{ 
                   \NUM  & \SG \\
                   \PRED & \Sandy } \\
     \PRED & \snore }
\end{center}
\caption{\label{f:cf1} 
  The c-structure and f-structure for \phon{Sandy snores} generated
  by the fragment (\ref{e:gr1a}--\ref{e:gr1e}).}
\end{figure}

\noindent
In this method, both agreeing items constrain the shared f-structure
element using defining equations.  For example, the 
grammar fragment in (\ref{e:gr1a}--\ref{e:gr1e})
generates exactly the two sentences in (\ref{e:agr1}).  The
c-structure and f-structure generated by this fragment for
(\ref{e:agr1a}) is depicted in Figure~\ref{f:cf1}.

\begin{eqnarray}
 \label{e:gr1a} \phon{Sandy} & \NP &
 \begin{array}[t]{l} (\up \PRED) = \Sandy\\ (\up \NUM) = \SG \end{array} \\
 \label{e:gr1b} \phon{Professors} & \NP &
 \begin{array}[t]{l} (\up \PRED) = \professor\\ (\up \NUM) = \PL \end{array} \\
 \label{e:gr1c} \phon{snores} & \VP &
 \begin{array}[t]{l} \up \PRED) = \snore\\ \underline{(\up \SUBJ\;\NUM) = \SG}
 \end{array}\\
 \label{e:gr1d} \phon{snore} & \VP &
 \begin{array}[t]{l} \up \PRED) = \snore\\ \underline{(\up \SUBJ\;\NUM) = \PL}
 \end{array}
\end{eqnarray}
\begin{equation} \label{e:gr1e}
 \S \; \longrightarrow \;
  \begin{array}[t]{c} \NP \\ (\up\SUBJ) = \down \end{array} 
  \begin{array}[t]{c} \VP \\ \up = \down \end{array}
\end{equation}

The lexical entries for subject $\NP$s require that the value of their
$\NUM$ attribute is $\SG$ or $\PL$ as appropriate.  In addition, the
underlined equation in each verb's lexical entry also requires that
this value is appropriate for the verb's inflection.  If the subject
and the verb require different values for this f-structure element (as
in the ungrammatical \phon{*Professors snores}), the corresponding
f-description will require this element to be equal to two different
values (e.g., $\SG$ and $\PL$).  However, the well-formedness
conditions on f-structures do not permit this
\cite{Kaplan82,Johnson95b} so the f-descriptions associated with such
sentences are inconsistent, and the sentences themselves are correctly
predicted to be ungrammatical.  

Thus this method functions by arranging for ungrammatical sentences to
be associated with an inconsistent f-description.  This observation is
in fact quite general: if all grammatical relationships are described
using defining equations (i.e., if we restrict
attention to the monotonic constraints) then the only way such
an equation can have a grammatical ``effect'' is by being inconsistent
with other equations, i.e., by ``causing'' ungrammaticality.

More precisely, suppose we identify a subset of the elements of a
f-structure as follows.  The {\em semantically interpreted elements}
are those which serve as the input to the semantic interpretation
procedure (in the framework of Dalrymple et.\ al.\ these elements
are associated with glue language formulae at some stage during the
interpretation process).  The idea is the semantically uninterpreted
elements can be deleted from an f-structure without changing its
semantic interpretation.  In a typical LFG, the values of attributes
such as $\PRED$, $\SUBJ$, $\OBJ$, etc., are semantically interpreted,
while the values of $\CASE$ and $\GENDER$ (in a grammatical gender
language) are not semantically interpreted.

Now consider a ``pure unification'' grammar without non-monotonic
devices such as ``constraining equations'', e.g., in which all
equations are defining equations, such as the PATR grammars of
\namecite{Shieber86}.  These are grammars in which all linguistic
relationships are expressed with defining equations.  It is
possible to show that in such a grammar, if an equation which equates
only non-semantic values is not inconsistent with other equations on
some input, then deleting it from the grammar does not affect the
language generated or the interpretations assigned.  (A similiar
observation holds in monotonic grammars such as HSPG).

This means that if all grammatical relationships are described using
defining equations, a
nonsemantic feature defining equation only has an effect on
the language generated if somewhere else in the
grammar there are defining equations that are inconsistent with this
one.
For example, there is no point in adding a defining equation that
introduces an attribute that does not appear elsewhere in the grammar,
such as
\begin{equation} \label{e:romance}
     (\up \att{HISTORICAL-ORIGIN}) = \att{ROMANCE}
\end{equation}
unless other defining equations that can possibly be inconsistent
with it are also introduced.  But in order to be inconsistent
with (\ref{e:romance}) these other equations must require the
attribute's value to be {\em different} to the value specified
in the former equation, e.g.,
\[
     (\up \att{HISTORICAL-ORIGIN}) = \att{GERMANIC}.
\]

Thus with defining equations alone, different grammatical properties
are based on feature {\em oppositions} or constrasts.  The formal
machinery of these monotonic ``pure unification'' grammars does not
completely support non-constrastive or ``privative'' feature values.

Indeed, f-structures seem to have been specifically designed to enable
systems of defining equations to be inconsistent.  For example, if we
removed either the ``functionality'' axiom (which requires attributes
to be single-valued) or the ``constant-constant'' clash axiom (which
specifies that distinct constants denote distinct f-structure
elements) from the formal definition of f-structures, then
f-descriptions such as
\[
	(f\; \CASE) = \ACC, (f\;\CASE) = \DAT
\]
would not be inconsistent.  R-LFG does not possess either the
functionality axiom or the constant-constant clash axiom,
and hence it does permit a single constituent to bear two
such distinct features, so long as both are checked or consumed
as described below.

\subsection{Agreement using defining and constraining equations}

\noindent
Writers of LFGs often employ constraining equations in order
to describe asymmetric linguistic relationships.  The subject-verb
agreement examples (\ref{e:agr1}) would be described using this method
by replacing the lexical entries (\ref{e:gr1c}--\ref{e:gr1d}) with
the following. 
\begin{eqnarray}
 \label{e:gr2c} \phon{snores} & \VP &
 \begin{array}[t]{l} \up \PRED) = \snore\\ \underline{(\up \SUBJ\;\NUM) \eqc \SG}
 \end{array}\\
 \label{e:gr2d} \phon{snore} & \VP &
 \begin{array}[t]{l} \up \PRED) = \snore\\ \underline{(\up \SUBJ\;\NUM) \eqc \PL}
 \end{array}
\end{eqnarray}
These entries differ from the previous ones in that the underlined
defining equations have been replaced with constraining equations.

While these two fragments both generate the same language in this
case, in general the two methods for describing agreement behave quite
differently.  For example, if an \NP's f-description contains the
constraint equation
\begin{equation} \label{e:constraint}
  (\up \CASE) \eqc \ACC 
\end{equation}
then this \NP\ must be independently ``assigned'' a value for the Case
feature in order for the f-structure to be well-formed.

This method behaves quite differently to the method that only uses
defining equations.  It does not rely on feature oppositions in the
same way that the defining equation method does.  For example, the
constraint equation (\ref{e:constraint}) requiring that the $\NP$
receive an $\ACC$ case value does not rely on the existence of other
Case values besides $\ACC$; it functions just as well if $\ACC$ is the
only Case value used in the grammar.  That is, while a defining
equation ensures that an attribute has one value rather than another,
a constraining equation ensures in addition that the feature has in
fact been given a value independently.  Thus this method more fully
supports privative features than the defining equation method does.

Further, the constraining equation method does not rely on the
functionality axiom or the constant-constant clash axioms in the same
way that the defining equation method does.  For example, even if
the functionality requirement on f-structures were relaxed so that the
defining equations in the f-description for (\ref{e:agr1a}) could have
the second minimal f-structure solution depicted in
Figure~\ref{f:nonfunct} besides the one depicted in
Figure~\ref{f:cf1}, that f-structure would fail to satisfy the
constraining equation expressing subject-verb agreement, and
so would be ill-formed for independent reasons.

\begin{figure}
\begin{center}
\fs{\SUBJ & \tfs{ \PRED & \Sandy \\
                  \NUM  & \SG } \\
    \SUBJ & \tfs{ \NUM   & $\kern-1.5ex\eqc$ \SG } \\
    \PRED & \snore }
\end{center}
\caption{A alternative minimal f-structure solution to the
 f-description for (\protect\ref{e:agr1a}) obtained by relaxing
 the functionality requirements on f-structures.  Note that this
 f-structure never the less does not satisfy the constraining
 equations expressing subject-verb agreement. \label{f:nonfunct}}
\end{figure}


In fact, feature structures in R-LFG behave very much in this way.
While attributes are permitted to be single-valued, no feature
structure axiom forces them to be so.  But since grammatical relationships
are described in a way very similiar to the constraining equation method,
in general the grammatical requirements of predicates will require
that attributes are single-valued.

\subsection{Resource management in LFG}

\noindent
The constraining equation method of describing agreement relationships
can be described in terms of {\em resources}, where the resource is
the feature value of the shared f-structure entity.  Each such feature
value is is {\em produced} by {\em one or more} defining equations,
and is {\em consumed} by {\em zero or more} constraining equations.
This pattern of resource management is formalized by Intuitionistic
Logic.

Interestingly, the special properties LFG endows the values of
$\PRED$ attributes with provides them with special resource
management properties also.  The values of $\PRED$ attributes
must be {\em produced} by {\em exactly one} argument, and must be 
{\em consumed} by {\em one or more} predicates.  The logic
LPC developed by \namecite{vanBenthem95}
formalizes this resource management.

Thus LFG already incorporates a number of mechanisms which can be seen
as performing resource management.  R-LFG attempts to describe all
syntactic relationships in terms of such resource management.  Identifying
the appropriate resource management mode for a particular grammatical
relationship is a key step in developing its R-LFG description.

\section{Resource accounting in R-LFG}

\noindent
\namecite{Johnson97} formally defines R-LFG's f-terms and presents a
logic that describes the resource management relationships between
features; that paper should be consulted for full details.

Informally, a f-term is a configuration of one or resources.
A resource is identified by its {\em type}.  The types 
$e$ and $t$ identify semantically contentful entities
(these are the types of individuals and truth values respectively),
while types such as $\NOM$ and $\ACC$ identify semantically
vacuous entities (which are interpreted by constants, and whose
value is systematically ignored by any function that takes them
as an argument).

F-terms describe recursive structures of resources.  Types are
f-terms, and if $\alpha_1, \ldots, \alpha_n$ are f-terms then:
\begin{description}
\item[$\alpha, \ldots, \alpha_n$] is the {\em multiset} of 
 resource structures $\{ \alpha_1, \ldots, \alpha_n \}$ (order is 
 unimportant in a multiset,
 but the number of times an element appears is important),
\item[$f\;\alpha$] is the result of {\em embedding} the structure 
 $\alpha$ under the attribute $f$,
\item[$f_1 \ldots f_m = g_1 \ldots g_n$] is a path equation which 
 {\em restructures} an f-term
 by moving a resource structure embedded under the sequence of
 attributes $f_1 \ldots f_m$ so that it is located under the sequence
 of attributes $g_1 \ldots g_n$,
\item[$\alpha_1 \limp \alpha_2$] is a {\em linear implication}, i.e.,
 a structure which consumes an $\alpha_1$ in order to produce
 an $\alpha_2$, and
\item[$\opt{\alpha}$] is an {\em optional} occurence of the structure
 $\alpha$.
\end{description}

An f-term describes a structure of resources.  The f-term associated
with a sentence is required to
simplify to a single resource of type $t$ in order for the sentence to
be grammatical.  (This single requirement subsumes both the requirement
that the f-description be satisfiable and the requirement that the
Linear Logic glue formula simplify to an expression of type $t$ in
standard LFG).
An f-term simplifies by applying linear implications,
restructuring using path equations, distributing attributes over
multisets, and either deleting optional elements or replacing them
with their non-optional counterpart.

Attributes are permitted, but not required, to distribute and factor 
over multisets.
That is, the following biimplication holds, where $f$ is an attribute
and $\alpha_1$ and $\alpha_2$ are f-terms:
\[
 f(\alpha_1 , \alpha_2) \;\Leftrightarrow\; (f\,\alpha_1), (f\,\alpha_2).
\]
Unlike LFG, R-LFG does not require that attributes are single-valued,
nor does it enforce a constant-constant clash.  Every f-term is
``satisfiable'' in that it represents some configuration of resources;
grammaticality is determined by whether those resources can combine
to produce a single element of type $t$ (the type of a saturated
proposition).

\subsection{Nominative Case marking in English}

\noindent
A simple R-LFG fragment which describes structural nominative case
assignment to subject $\NP$s is presented below.  The lexical entry
for the pronoun \phon{she} in (\ref{e:she}) requires it to consume a
$\NOM$ case resource in order to produce a resource of type $e$, and
the lexical entry for the verb \phon{snores} in (\ref{e:snores})
requires it to consume a resource of type $e$ embedded within a
$\SUBJ$ attribute in order to produce a resource of type $t$.  The
syntactic rule (\ref{e:SNPVP}) specifies how the f-terms associated
with the $\NP$ and $\VP$ (referred to by the meta-variable `$\down$' just
as in LFG) are to be combined to produce the f-term for the $\S$.  In
this case, a multiset consisting of the $\NP$'s f-term and a $\NOM$
case resource is embedded within a $\SUBJ$ attribute, which together
with the f-term associated with the $\VP$ yields the multiset
f-term associated with the $\S$.

\begin{eqnarray}
\hbox to 0.7in{\phon{she}} & \NP & \NOM \limp e \label{e:she} \\
\hbox to 0.7in{\phon{snores}} & \VP & \SUBJ\; e \limp t \label{e:snores}
\end{eqnarray}
\begin{equation} \label{e:SNPVP}
\S \;\longrightarrow\; 
  \begin{array}[t]{c} \NP\\ \SUBJ(\NOM, \down) \end{array} \; 
  \begin{array}[t]{c} \VP\\ \down \end{array} 
\end{equation}

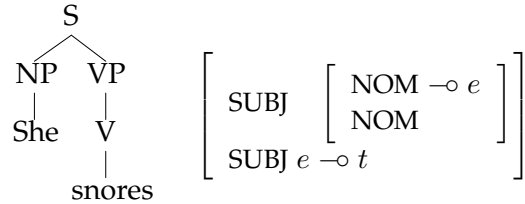
\begin{figure}
\begin{center}
\mbox{\lower3em\hbox{
\begin{picture}(54,78)(0,-78)
%
\put(24,-8){S}
\drawline(27,-12)(13,-22)
\put(6,-30){NP}
\drawline(13,-34)(13,-44)
\put(5,-52){She}
\drawline(27,-12)(40,-22)
\put(33,-30){VP}
\drawline(40,-34)(40,-44)
\put(36,-52){V}
\drawline(40,-56)(40,-66)
\put(27,-74){snores}
\end{picture}}}
\hspace{0.2in}
\fs{ \SUBJ & \tfs{ 
                   \noatt{$\NOM  \limp e$} \\
                   \noatt{\NOM} } \\
     \noatt{$\SUBJ\; e \limp t$}  }
\end{center}
\caption{\label{f:cf2} 
  The c-structure and f-term for \phon{Sandy snores} generated
  by the fragment (\ref{e:she}--\ref{e:SNPVP}).  The f-term simplifies
  straightforwardly to type $t$. }
\end{figure}

This fragment generates the c-structure and f-term depicted in
Figure~\ref{f:cf2}.  The f-term simplifies to type $t$ in the following
steps:
\begin{eqnarray}
 & \SUBJ(\NOM, \NOM \limp e),\; \SUBJ\; e \limp t &  \\
 & \SUBJ\; e, \SUBJ\; e \limp t & \\
 & t & 
\end{eqnarray}

\subsection{Icelandic Quirky Case Marking}

\def\drengurinn{\gp{\em drengurinn}{the-boy.\mbox{nom}}}
\def\drengina{\gp{\em drengina}{the-boys.\mbox{acc}}}
\def\kyssti{\gp{\em kyssti}{kissed}}
\def\vantar{\gp{\em vantar}{lacks}}
\def\stulkuna{\gp{\em st\'{u}lkuna}{the-girl.\mbox{acc}}}
\def\mat{\gp{\em mat}{food.\mbox{acc}}}
\def\hann{\gp{\em hann}{he.\mbox{nom}}}
\def\virdist{\gp{\em vir{\eth}ist}{seems}}
\def\elska{\gp{\em elska}{love}}
\def\hana{\gp{\em hana}{her.\mbox{acc}}}
\def\peninga{\gp{\em peninga}{money.\mbox{acc}}}
\def\vanta{\gp{\em vanta}{lack}}

\noindent
Quirky Case marking in Icelandic presents a more complex
array of linguistic data which exercises a wider range of f-term
machinery.  The analysis presented here is based heavily on
the LFG analysis of \namecite{Andrews82}.

In Icelandic, subject $\NP$s are usually case marked nominative,
as in (\ref{e:icenorm}).  However, a few verbs, such
as \phon{vantar} `lacks' exceptionally case mark their
subject $\NP$s with accusative or some other non-nominative ``quirky''
case (\ref{e:icequirky}).  The subjects of subject raising verbs, such
as \phon{vir{\eth}ist} `seems', usually appear in nominative
case (\ref{e:iceraising}), but if the embedded verb is 
a quirky case assigning verb then the matrix subject is
assigned the quirky case, rather than nominative (\ref{e:iceqraising}).

\eenumsentence{
\item \mbox{\drengurinn\ \kyssti\ \stulkuna} \  \label{e:icenorm}\\
      \strut `The boy kissed the girl'
\item \mbox{\drengina\ \vantar\ \mat} \   \label{e:icequirky}\\
      \strut `The boys lack food'
\item \mbox{\hann\ \virdist\ \elska\ \hana } \ \label{e:iceraising}\\
      \strut `He seems to love her'
\item \mbox{\hana\ \virdist\ \vanta\ \peninga}\ \label{e:iceqraising}\\
      \strut `She seems to lack money'
}

This pattern of data receives a straightforward informal account
in terms of case assignment if we make the following assumptions:
\begin{itemize}
\item All NPs must receive exactly one case,
\item Quirky case marking verbs always assign a quirky case,
\item Case is preserved in Raising and other grammatical operations, and
\item Structural nominative case is only optionally assigned.
\end{itemize}

Thus if a subject $\NP$ receives a quirky case, then that must
be the case that it appears in.  On the other hand, if the
subject $\NP$ is not assigned a quirky case, then the only case
available is structural nominative case.

This account can be formalized in R-LFG as follows.  The
phrase structure rules for this Icelandic fragment are the following.
\begin{equation} \label{e:iceSNPVP}
 \S \;\longrightarrow\;
   \tcarray{ \NP \\ \SUBJ(\opt{\NOM}, \down) }
   \tcarray{ \VP \\ \down }
\end{equation}
\begin{equation} \label{e:iceVP}
 \VP \;\longrightarrow\;
   \tcarray{ \V \\ \down }
   \left( \tcarray{ \NP \\ \OBJ\;\down } \right)
   \left( \tcarray{ \VP \\ \XCOMP\;\down } \right)
\end{equation}
The phrase structure rule (\ref{e:iceSNPVP}) differs from the
corresponding English rule (\ref{e:SNPVP}) in that it optionally
embedds a $\NOM$ case under the $\SUBJ$ attribute.  The phrase
structure rule (\ref{e:iceVP}) introduces a verb, an optional direct
object $\NP$ and an optional $\VP$.  It embedds the direct object
$\NP$'s f-term under the $\OBJ$ attribute and the $\VP$'s f-term under
the $\XCOMP$ attribute, as is standard in LFG.

\begin{figure}
\def\NP{\mbox{$e$}}
\def\S{\mbox{$t$}}
\begin{center}\setlength{\unitlength}{0.00083333in}
\begingroup\makeatletter\ifx\SetFigFont\undefined%
\gdef\SetFigFont#1#2#3#4#5{%
  \reset@font\fontsize{#1}{#2pt}%
  \fontfamily{#3}\fontseries{#4}\fontshape{#5}%
  \selectfont}%
\fi\endgroup%
{\renewcommand{\dashlinestretch}{30}
\begin{picture}(5319,1389)(0,-10)
\path(3595,1287)(3445,1287)(3445,837)(3595,837)
\path(4420,1287)(4570,1287)(4570,837)(4420,837)
\path(3595,537)(3445,537)(3445,87)(3595,87)
\path(4420,537)(4570,537)(4570,87)(4420,87)
\path(370,762)(370,237)
\path(1270,462)(1270,237)
\path(2170,462)(2170,312)
\path(1270,687)(1720,837)(2170,687)
\path(370,987)(1045,1137)(1720,987)
\path(2995,1362)(2845,1362)(2845,12)(2995,12)
\path(4870,1362)(5020,1362)(5020,12)(4870,12)
\put(370,87){\makebox(0,0)[b]{\smash{{{\SetFigFont{10}{12.0}{\familydefault}{\mddefault}{\updefault}\drengurinn}}}}}
\put(1270,87){\makebox(0,0)[b]{\smash{{{\SetFigFont{10}{12.0}{\familydefault}{\mddefault}{\updefault}\kyssti}}}}}
\put(2170,87){\makebox(0,0)[b]{\smash{{{\SetFigFont{10}{12.0}{\familydefault}{\mddefault}{\updefault}\stulkuna}}}}}
\put(1270,537){\makebox(0,0)[b]{\smash{{{\SetFigFont{10}{12.0}{\familydefault}{\mddefault}{\updefault}V}}}}}
\put(2170,537){\makebox(0,0)[b]{\smash{{{\SetFigFont{10}{12.0}{\familydefault}{\mddefault}{\updefault}NP}}}}}
\put(1720,837){\makebox(0,0)[b]{\smash{{{\SetFigFont{10}{12.0}{\familydefault}{\mddefault}{\updefault}VP}}}}}
\put(370,837){\makebox(0,0)[b]{\smash{{{\SetFigFont{10}{12.0}{\familydefault}{\mddefault}{\updefault}NP}}}}}
\put(1045,1137){\makebox(0,0)[b]{\smash{{{\SetFigFont{10}{12.0}{\familydefault}{\mddefault}{\updefault}S}}}}}
\put(2920,987){\makebox(0,0)[lb]{\smash{{{\SetFigFont{10}{12.0}{\familydefault}{\mddefault}{\updefault}$\SUBJ$}}}}}
\put(3520,1137){\makebox(0,0)[lb]{\smash{{{\SetFigFont{10}{12.0}{\familydefault}{\mddefault}{\updefault}$\opt{\NOM}$}}}}}
\put(3520,912){\makebox(0,0)[lb]{\smash{{{\SetFigFont{10}{12.0}{\familydefault}{\mddefault}{\updefault}$\NOM\limp\NP$}}}}}
\put(3520,387){\makebox(0,0)[lb]{\smash{{{\SetFigFont{10}{12.0}{\familydefault}{\mddefault}{\updefault}$\ACC$}}}}}
\put(3520,162){\makebox(0,0)[lb]{\smash{{{\SetFigFont{10}{12.0}{\familydefault}{\mddefault}{\updefault}$\ACC\limp\NP$}}}}}
\put(2920,237){\makebox(0,0)[lb]{\smash{{{\SetFigFont{10}{12.0}{\familydefault}{\mddefault}{\updefault}$\OBJ$}}}}}
\put(2920,612){\makebox(0,0)[lb]{\smash{{{\SetFigFont{10}{12.0}{\familydefault}{\mddefault}{\updefault}$\OBJ\;\NP\limp\SUBJ\;\NP\limp\S$}}}}}
\end{picture}
}\end{center}
\caption{\label{f:ice-simple}
  The c-structure and f-term for the single clause non-quirky Icelandic example (\ref{e:icenorm}) generated by (\ref{e:iceSNPVP}--\ref{e:ices1c}).}
\end{figure}
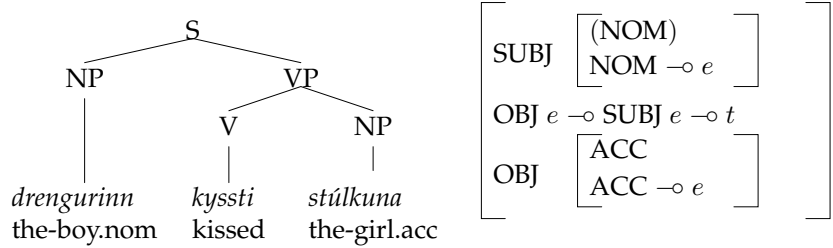

The lexical entries (\ref{e:ices1a}--\ref{e:ices1c}) are required to generate 
the non-quirky single clause example (\ref{e:icenorm}).  The c-structure
and f-term associated with this example are shown in Figure~\ref{f:ice-simple}.
It is straightforward to check that this f-term reduces to $t$.

\begin{eqnarray}
\hbox to 0.7in{\mbox{\em drengurinn}} & \NP & \NOM\limp e \label{e:ices1a}\\ 
\hbox to 0.7in{\mbox{\em st\'{u}lkuna}} & \NP & \ACC \limp e \\ 
\hbox to 0.7in{\mbox{\em kyssti}} & \V & \OBJ\; e \limp \SUBJ\; e \limp t, \OBJ\;\ACC \label{e:ices1c}
\end{eqnarray}

The single clause quirky case marked example is only slightly more
complex.  It can be described with the two additional lexical entries
(\ref{e:ices2a}--\ref{e:ices2b}).

\begin{eqnarray}
\hbox to 0.6in{\mbox{\em drengina}} & \NP & \ACC \limp e \label{e:ices2a} \\
\hbox to 0.6in{\mbox{\em vantar}} & \V & \OBJ\; e \limp \SUBJ\; e \limp t, \OBJ\;\ACC, \ul{\SUBJ\;\ACC} \label{e:ices2b}
\end{eqnarray}

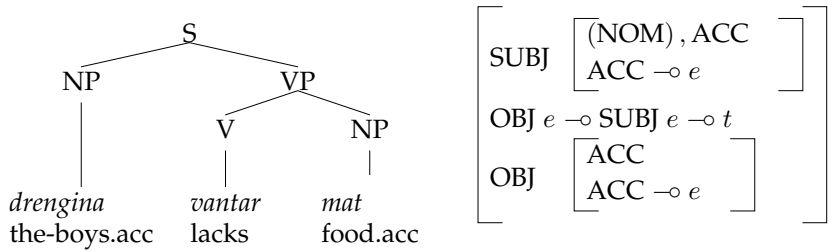
\begin{figure}
\def\NP{\mbox{$e$}}
\def\S{\mbox{$t$}}
\begin{center}\setlength{\unitlength}{0.00083333in}
\begingroup\makeatletter\ifx\SetFigFont\undefined%
\gdef\SetFigFont#1#2#3#4#5{%
  \reset@font\fontsize{#1}{#2pt}%
  \fontfamily{#3}\fontseries{#4}\fontshape{#5}%
  \selectfont}%
\fi\endgroup%
{\renewcommand{\dashlinestretch}{30}
\begin{picture}(5242,1389)(0,-10)
\path(3518,537)(3368,537)(3368,87)(3518,87)
\path(4343,537)(4493,537)(4493,87)(4343,87)
\path(293,762)(293,237)
\path(1193,462)(1193,237)
\path(2093,462)(2093,312)
\path(1193,687)(1643,837)(2093,687)
\path(293,987)(968,1137)(1643,987)
\path(2918,1362)(2768,1362)(2768,12)(2918,12)
\path(4793,1362)(4943,1362)(4943,12)(4793,12)
\path(3518,1287)(3368,1287)(3368,837)(3518,837)
\path(4643,1287)(4793,1287)(4793,837)(4643,837)
\put(293,87){\makebox(0,0)[b]{\smash{{{\SetFigFont{10}{12.0}{\familydefault}{\mddefault}{\updefault}\drengina}}}}}
\put(1193,87){\makebox(0,0)[b]{\smash{{{\SetFigFont{10}{12.0}{\familydefault}{\mddefault}{\updefault}\vantar}}}}}
\put(2093,87){\makebox(0,0)[b]{\smash{{{\SetFigFont{10}{12.0}{\familydefault}{\mddefault}{\updefault}\mat}}}}}
\put(1193,537){\makebox(0,0)[b]{\smash{{{\SetFigFont{10}{12.0}{\familydefault}{\mddefault}{\updefault}V}}}}}
\put(2093,537){\makebox(0,0)[b]{\smash{{{\SetFigFont{10}{12.0}{\familydefault}{\mddefault}{\updefault}NP}}}}}
\put(1643,837){\makebox(0,0)[b]{\smash{{{\SetFigFont{10}{12.0}{\familydefault}{\mddefault}{\updefault}VP}}}}}
\put(293,837){\makebox(0,0)[b]{\smash{{{\SetFigFont{10}{12.0}{\familydefault}{\mddefault}{\updefault}NP}}}}}
\put(968,1137){\makebox(0,0)[b]{\smash{{{\SetFigFont{10}{12.0}{\familydefault}{\mddefault}{\updefault}S}}}}}
\put(2843,987){\makebox(0,0)[lb]{\smash{{{\SetFigFont{10}{12.0}{\familydefault}{\mddefault}{\updefault}$\SUBJ$}}}}}
\put(3443,1137){\makebox(0,0)[lb]{\smash{{{\SetFigFont{10}{12.0}{\familydefault}{\mddefault}{\updefault}$\opt{\NOM},\ACC$}}}}}
\put(3443,912){\makebox(0,0)[lb]{\smash{{{\SetFigFont{10}{12.0}{\familydefault}{\mddefault}{\updefault}$\ACC\limp\NP$}}}}}
\put(3443,387){\makebox(0,0)[lb]{\smash{{{\SetFigFont{10}{12.0}{\familydefault}{\mddefault}{\updefault}$\ACC$}}}}}
\put(3443,162){\makebox(0,0)[lb]{\smash{{{\SetFigFont{10}{12.0}{\familydefault}{\mddefault}{\updefault}$\ACC\limp\NP$}}}}}
\put(2843,237){\makebox(0,0)[lb]{\smash{{{\SetFigFont{10}{12.0}{\familydefault}{\mddefault}{\updefault}$\OBJ$}}}}}
\put(2843,612){\makebox(0,0)[lb]{\smash{{{\SetFigFont{10}{12.0}{\familydefault}{\mddefault}{\updefault}$\OBJ\;\NP\limp\SUBJ\;\NP\limp\S$}}}}}
\end{picture}
}\end{center}
\caption{\label{f:ice-qsimple}
  The c-structure and f-term for the single clause quirky case
  example (\ref{e:icequirky}) generated 
  by (\ref{e:iceSNPVP}--\ref{e:ices2b}).}
\end{figure}

The lexical entry for the quirky case marking verb \phon{vantar}
`lacks' in (\ref{e:ices2b}) differs from that for the non-quirky verb
\phon{kyssti} `kissed' in that it assigns an accusative case to its
subject (in the underlined part of the f-term)
as well as to its object.  The c-structure and f-term for
(\ref{e:icequirky}) are depicted in Figure~\ref{f:ice-qsimple}.
Again, it is straightforward to check that the f-term reduces to $t$.
Note that if the subject were replaced with a nominative $\NP$ the
f-term would no longer reduce to $t$, since the $\ACC$ case feature
embedded under the $\SUBJ$ attribute could not be consumed. 

\begin{figure}
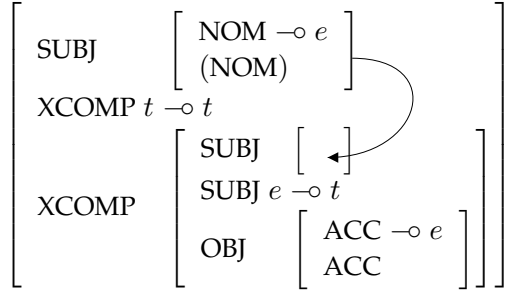

\begin{center}
\fs{ \SUBJ & \tfs{ \noatt{$\NOM\limp e$} \\
                  \noatt{$\opt{\NOM}$} } \nodepoint{a} \\
     \noatt{$\XCOMP\; t \limp t$} \\
     \XCOMP & \bfs{ \SUBJ & \fs{\ \nodepoint{b}\ } \\
                   \noatt{$\SUBJ\; e \limp t$} \\
		   \OBJ & \bfs{\noatt{$\ACC\limp e$} \\
                              \noatt{$\ACC$} } } }
\anodecurve[r]{a}[r]{b}{0.5in}
\end{center}
\caption{\label{f:iceraising}
  The f-term for the non-quirky Subject Raising
  example (\ref{e:iceraising}) generated 
  by (\ref{e:iceSNPVP}--\ref{e:ices}).}
\end{figure}

The formalization of the non-quirky case Subject Raising example
(\ref{e:iceraising}) is very similiar to the standard LFG account
of Subject Raising \cite{Bresnan82}.  
The lexical entry (\ref{e:icerv})
for the Raising verb \phon{vir{\eth}ist} `seems' contains the path
equation $\SUBJ = \XCOMP\;\SUBJ$ which permits resources embedded 
under the $\SUBJ$ attribute to be restructured under the $\XCOMP\;\SUBJ$
attributes.  In this example, a resource of type $e$ is lowered
into the embedded clause.  The f-term associated with this example
is depicted in Figure~\ref{f:iceraising}.  It is straightforward
to check that this reduces to $t$.

\begin{eqnarray}
\hbox to 0.6in{\mbox{\em vir{\eth}ist}} & \V & \XCOMP\; t \limp t, \SUBJ = \XCOMP\;\SUBJ \label{e:icerv} \\
\hbox to 0.6in{\mbox{\em elska}} & \V & \OBJ\; e \limp \SUBJ\; e \limp t, \OBJ\;\ACC \label{e:ices}
\end{eqnarray}

\begin{figure}
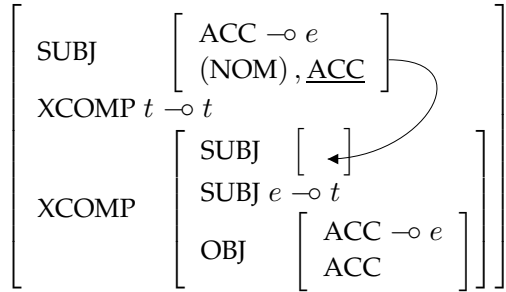

\begin{center}
\fs{ \SUBJ & \tfs{ \noatt{$\ACC\limp e$} \\
                  \noatt{$\opt{\NOM},\ul{\ACC}$} } \nodepoint{a} \\
     \noatt{$\XCOMP\; t \limp t$} \\
     \XCOMP & \bfs{ \SUBJ & \fs{\ \nodepoint{b}\ } \\
                   \noatt{$\SUBJ\; e \limp t$} \\
		   \OBJ & \bfs{\noatt{$\ACC\limp e$} \\
                              \noatt{$\ACC$} } } }
\anodecurve[r]{a}[r]{b}{0.5in}
\end{center}
\caption{\label{f:iceqraising}
  The f-term for the quirky case marked Subject Raising
  example (\ref{e:iceqraising}) generated 
  by (\ref{e:iceSNPVP}--\ref{e:ices}).}
\end{figure}

The syntactic rules and lexical entries introduced above
that are independently needed to account for quirky case marking
in single clause constructions and for Subject Raising without
quirky case also correctly account for the interaction of those two
constructions, which was presented in (\ref{e:iceqraising}) on
page~\pageref{e:iceqraising}.  The f-term for this
example is shown in Figure~\ref{f:iceqraising}.

Just as in the single clause quirky case marking example (\ref{e:icequirky}),
the subject $\NP$ is assigned both an accusative case and an optional
nominative case, so only an accusative subject $\NP$ can appear.

\section{Conclusion}

\noindent
This paper has introduced a simplified version of LFG called R-LFG
in which a single representation called an f-term plays the role
of both f-description and f-structure.  LFG's f-structure well-formedness
constraints are re-expressed in terms of feature resource dependencies,
which permits them to be checked by the same mechanism that performs
semantic interpretation. It is not implausible that this can be done
for many, if not most, LFG analyses,
as many standard LFG analyses already have a resource oriented character,
and it seems that the ``core'' LFG analyses of Raising, Control, etc.,
can be straightforwardly reexpressed in R-LFG.

Even if it turns out that the R-LFG project is ultimately untenable---
perhaps it will be possible to demonstrate that some linguistically
necessary properties of f-structures simply cannot be adequately
captured using the resource logic machinery utilized for semantic
interpretation---this research may still contribute by providing
an alternative perspective on feature interactions in grammar
and suggesting
modifications or extensions to the standard LFG framework.

\bibliography{mj}
\end{document}